\pdfoutput=1

\documentclass[11pt]{article}

\usepackage[]{acl}

\usepackage{times}
\usepackage{latexsym}

\usepackage[T1]{fontenc}

\usepackage[utf8]{inputenc}

\usepackage{microtype}

\usepackage{graphicx}
\usepackage{booktabs}
\usepackage{adjustbox}
\usepackage{multicol}
\usepackage{multirow}
\usepackage{lipsum}

\title{
Benchmarking Answer Verification Methods for Question Answering-Based Summarization Evaluation Metrics}

\author{Daniel Deutsch and Dan Roth \\
  Department of Computer and Information Science \\
  University of Pennsylvania \\
  \texttt{\{ddeutsch,danroth\}@seas.upenn.edu} \\}
  
    \newif\ifcomments
    \commentstrue
    \ifcomments
        \providecommand\dd[1]{\textcolor{blue}{[DD: #1]}}
        \providecommand\dr[1]{\textcolor{green}{[DR: #1]}}
        \providecommand\todo[1]{\textcolor{red}{[TODO: #1]}}
    \else
        \providecommand{\dd}[1]{}
        \providecommand{\dr}[1]{}
        \providecommand{\todo}[1]{}
    \fi

\newcommand{\qaeval}{QA\-Eval}
\newcommand{\summeval}{Summ\-Eval}

\begin{document}
\maketitle

\begin{abstract}
Question answering-based summarization evaluation metrics must automatically determine whether the QA model's prediction is correct or not, a task known as answer verification.
In this work, we benchmark the lexical answer verification methods which have been used by current QA-based metrics as well as two more sophisticated text comparison methods, BERTScore and LERC.
We find that LERC out-performs the other methods in some settings while remaining statistically indistinguishable from lexical overlap in others.
However, our experiments reveal that improved verification performance does not necessarily translate to overall QA-based metric quality:
In some scenarios, using a worse verification method --- or using none at all --- has comparable performance to using the best verification method, a result that we attribute to properties of the datasets.\footnote{
    Our code is available at \url{http://cogcomp.org/page/publication_view/966}.
}
\end{abstract}
\section{Introduction}
A recent trend in summarization metrics is evaluating the quality of a summary via question answering \citep[QA;][]{EyalBaEl19,SLPS19,SDLPSWG21,DurmusHeDi20,WangChLe20,DeutschBeRo21}.
These metrics compare the semantic content of two texts (e.g., the reference and candidate summaries) by generating questions from one and answering those questions against the other.
The amount of common semantic content is proportional to the number of questions which are answered correctly.

\begin{figure}
    \centering
    \includegraphics[width=0.95\columnwidth]{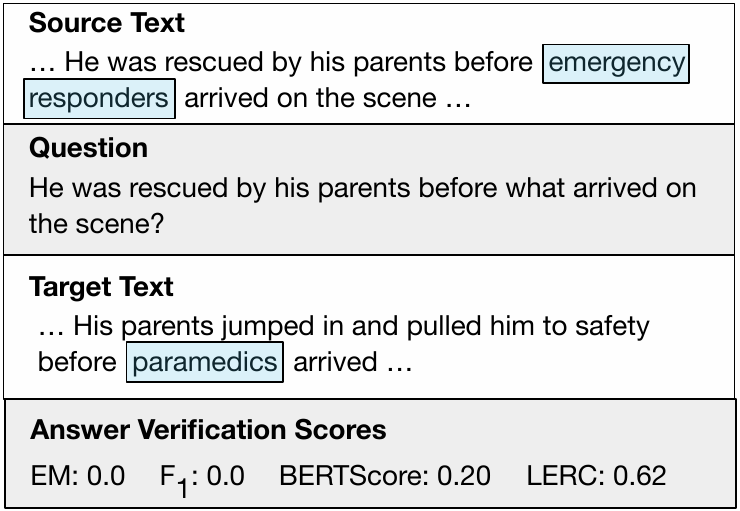}
    \caption{In the answer verification task, the metrics score how likely two phrases from different contexts have the same meaning.
    Here, the metrics at the bottom score the similarity between ``emergency responders,'' which was used to generate the question from the source text, and ``paramedics,'' the predicted answer from a QA model in the target text.}
    \label{fig:intro_figure}
\end{figure}

A critical step of QA-based evaluation metrics is to verify whether the QA model's prediction is correct, a task known as answer verification (see Fig.~\ref{fig:intro_figure}).
This helps to both suppress noisy output from the QA model as well as identify inconsistent information across the texts.

Answer verification is typically done by comparing the prediction to the expected answer by the exact match or token F$_1$ string comparison methods \citep{RZLL16}.
However, more sophisticated text comparison methods have been proposed in recent years, and it is unknown whether they provide a benefit in this particular scenario.

In this work, we benchmark various answer verification strategies for QA-based summarization evaluation metrics.
Our goal is to understand whether methods that are more advanced than lexical overlap are better able to classify phrases as having the same or different meaning as well as whether any such improvements result in the overall QA-based metric being better at replicating human judgments of summary quality.

We analyze four answer verification methods, exact match, token F$_1$, BERTScore \citep{ZKWWA20}, and LERC, \citep{CSSG20} in combination with two QA-based metrics,  QAEval \citep{DeutschBeRo21} and FEQA \citep{DurmusHeDi20}.

Based on a set of human annotations across two datasets, we find that LERC performs the best at the actual task of answer verification in general, although in some settings it is statistically indistinguishable from token F$_1$ (\S\ref{sec:answer_verif_perf}).
However, our results also show that any such improvement in verification performance does not always translate to a better QA-based evaluation metric (\S\ref{sec:end_to_end}).

We believe these results can be explained by properties of the QA metrics and the datasets.
When the QA model performance is high or the verification task is in some sense easy to do, it may not be necessary to have a sophisticated verification method or even use one at all.
Despite this, our recommendation is to use both token F$_1$ and LERC for answer verification since F$_1$ may suffice in some situations and we suspect LERC does provide additional benefits, although they are difficult to measure.
\section{Related Work \& Background}
\label{sec:related_work}
The majority of summarization evaluation metrics can be viewed as estimating how similar in meaning two pieces of text are.
For instance, ROUGE \citep{Lin04} does this by calculating the number of overlapping $n$-grams between the two texts.

Instead of directly comparing the entire texts, QA-based metrics identify specific phrases within the texts which should be compared, as follows.
First, a set of questions is automatically generated from one text.
Then, those questions are automatically answered against a second text to obtain a set of predicted answers.
The final score is proportional to the number of correct predictions, but determining whether those predictions are correct (the task of answer verification) is done by comparing the text of the prediction to the expected answer.
Therefore, instead of directly comparing the entire contents of the two texts, QA-based metrics instead reduce the scope of the problem to only comparing specific pairs of phrases.

Current QA-based metrics perform the answer verification step by lexical comparison, either exact match or token F$_1$.
Such metrics include \qaeval{} \citep{DeutschBeRo21}, FEQA \citep{DurmusHeDi20}, and more \citep{EyalBaEl19, WangChLe20, SLPS19,SDLPSWG21}.
However, any such function which calculates the similarity of arbitrary text can be used instead.
This includes embedding-based methods such as BERTScore \citep{ZKWWA20} or metrics which have been trained specifically to do this task, such as LERC \citep{CSSG20}.
Evaluating how these methods perform as answer verification methods for QA-based metrics compared to the lexical baselines is the scope of this work.

Other, related work has also benchmarked various answer verification methods \citep{CSSG19}, but do so as a method for evaluating QA performance rather than as part of a downstream task, as we do in this work.
Some concurrent work also tries to improve answer verification by expanding the set of possible expected answers via mining additional aliases from knowledge bases \citep{SiZhBo21}.
\section{Definitions \& Methods}
\label{sec:problem}
We define the answer verification task as the following:
Given a question, answer, the source text from which the QA pair was generated, a prediction, and the target text the prediction comes from, score how similar the meanings of the answer and prediction are (see Fig.~\ref{fig:intro_figure} for an example).\footnote{
    This is slightly different from the task defined by \citet{CSSG20} which does not include the source text because no such text exists in the standard definition of the reading comprehension task.
    However, we include it because the source text can be used to create a representation for the answer which may be better than using the question alone.
}
Answer verification is used by QA-based metrics to suppress noisy outputs from the QA model as well as identify when the QA prediction is correct with respect to the target text but incorrect with respect to the expected answer (e.g., unfaithful information).

We analyze four different answer verification methods.

\paragraph{Exact Match}
The exact match (EM) method compares the two phrases to see if they are identical (after light normalization).
EM assigns a score of 1 if the phrases are identical and 0 otherwise.

\paragraph{Token F$_1$}
The token F$_1$ comparison calculates an F$_1$ score based on the number of unigrams the two phrases have in common.
This is equivalent to the F$_1$ variant of ROUGE-1.

\paragraph{BERTScore}
BERTScore \citep{ZKWWA20} compares two pieces of text by aligning the texts' tokens according to which pairs have the highest BERT embedding cosine similarity.
We adapt BERTScore to answer verification by encoding the answer and prediction using their respective contexts, then calculating the BERTScore only between the two phrase encodings.
Since the output of BERTScore is often in a narrow range of values, we rescale the scores by defining 0 and 1 as the 2.5th and 97.5th percentiles of the BERTScores calculated over the whole dataset.
These changes were made to make the score more interpretable as well as prevent outliers from influencing the score rescaling.
In practice, we compute the BERTScore using embeddings obtained from RoBERTa-Large \citep{LOGDMJCLLZS19}.

\paragraph{LERC} \citet{CSSG20} proposed LERC, a learned metric for scoring how similar the expected and predicted answers to a question are conditioned on the question and the target text the prediction comes from.
The metric takes as input the target context, question, expected answer, and predicted answer and concatenates them into a single sequence separated by speical tokens.
Because it was designed for scoring reading comprehension predictions, it does not use the source text.
It then encodes the entire sequence with BERT and trains a regression layer on top of the encodings to predict a similarity score.
The learned metric was fine-tuned on 40k human annotations of how similar the two answers are on a scale from 1 to 5.
We rescale the output from LERC to be in the range $[0, 1]$.

\section{Experiments}
\label{sec:experiments}
The answer verification methods are evaluated independently (\S\ref{sec:answer_verif_perf}) as well as in combination with two QA-based metrics (\S\ref{sec:end_to_end}), \qaeval{} \citep{DeutschBeRo21} and FEQA \citep{DurmusHeDi20}.
\qaeval{} measures the content quality of a summary (does the summary contain ``summary-worthy'' information) by using a reference summary as the source text and candidate summary as the target text.
In contrast, FEQA estimates the faithfulness of the summary (does the summary contain information consistent with the input) by using the candidate summary as the source text and the input document as the target text.

The experiments are run on two datasets, TAC'08 \citep{DangOw08} and \summeval{} \citep{FKMSR21}.
These datasets have summaries generated by 58 and 16 models for 48 and 100 inputs, respectively, which are annotated with expert judgments.
Both \qaeval{} and FEQA are evaluated on \summeval{} because it contains annotations for both summary quality and faithfulness, whereas only \qaeval{} is evaluated on TAC'08 since it does not have faithfulness judgments.

\subsection{Answer Verification Performance}
\label{sec:answer_verif_perf}
First, we examine how well each answer verification method accurately scores manually labeled answer pairs from the summarization datasets.
For each QA metric and dataset combination, we ran the metric on the summaries, then randomly sampled 200 QA predictions (making 600 total).
Each prediction and expected answer were manually annotated by the authors for whether or not the two phrases share the same meaning.
See Appendix~\ref{sec:annotation_details} for additional details on the annotation procedure.

Ideally, the answer verification methods should both successfully classify phrases based on their meaning as well as provide a score close to 1 for phrases with the same meaning and close to 0 with different meanings.
These properties are quantified by the binary classification accuracy (assigning labels based on a threshold which maximizes this score) as well as the mean squared error (MSE) of the predicted scores, show in Table~\ref{tab:binary_acc}.

\begin{table}[t]
    \centering
    \begin{adjustbox}{width=\columnwidth}
    \begin{tabular}{lccccccc}
        \toprule
        \multirow{3.4}{*}{\textbf{Ans. Verif.}} & \multicolumn{4}{c}{\bf QAEval} & & \multicolumn{2}{c}{\bf FEQA} \\
         \cmidrule{2-5} \cmidrule{7-8}
         & \multicolumn{2}{c}{TAC'08} & \multicolumn{2}{c}{SummEval} & & \multicolumn{2}{c}{SummEval} \\
        & Acc & MSE & Acc & MSE & & Acc & MSE \\
        \midrule
        Majority Cls     & 51.5             & .49 & 78.5     & .22 & & 56.5 & .44 \\
        EM        & 64.5             & .36 & 78.5     & .46 & & 76.0 & .24 \\
        F$_1$     & \underline{84.0} & .19 & 79.5     & .25 & & \bf 91.0 & \underline{.10} \\
        BERTScore & \underline{81.0} & .16 & 79.5     & .20 & & 82.5 & .16 \\
        LERC      & \bf 85.0         & \bf .13 & \bf 88.0 & \bf .11 & & \underline{88.5} & \bf .09 \\
        \bottomrule
    \end{tabular}
    \end{adjustbox}
    \caption{The binary accuracies and mean squared errors of the answer verification methods evaluated on three metric-dataset combinations with 200 manually labeled examples each.
    Underlined values are statistically indistinguishable from those in bold under a single-tailed pairwise permutation test with $\alpha = 0.05$.
    }
    \label{tab:binary_acc}
\end{table}

We find that LERC is the only method with the best (or tied for the best) performance across all three metric-dataset combinations.
Despite LERC's significant improvement on the SummEval data with QAEval predictions, it is statistically indistinguishable from F$_1$ on the same dataset with FEQA predictions.
We believe this can be explained by which texts are being compared for each metric.
FEQA compares the generated summary to the input document.
Recent summarization models are known to copy heavily from the input with little high-level abstraction or rephrasing, so comparing phrases with token F$_1$ is likely to be quite successful.
In contrast, QAEval compares the reference and generated summaries.
The reference summaries are written by humans, and thus more likely to contain information from the input document which is expressed differently.
In such a scenario, the learned metric, LERC, shows strong improvements over F$_1$.

In general, we find that when BERTScore and LERC do improve over F$_1$, they do so by identifying paraphrases that have no tokens in common, which sometimes requires world knowledge.
Examples of this are included in Appendix~\ref{sec:examples}.

\subsection{Overall Metric Evaluation}
\label{sec:end_to_end}
Next, we investigate whether the differences in classification performance of the verification methods translate to downstream improvements in the overall quality of the QA-based metrics.
To do so, we evaluate different variants of the metrics that use each answer verification method.
For both QAEval and FEQA, the final score for the summary is the output of the answer verification method averaged over all of the QA pairs.\footnote{
    QAEval can also predict a question is unanswerable.
    In such cases, the score of the prediction is 0.
}

\paragraph{QAEval}
For QAEval, we report the standard system- and summary-level correlations of the metrics' scores to human judgments in Table~\ref{tab:qaeval_correlations} (due to space constraints, we refer the reader to \citet{DeutschDrRo21} for definitions of the correlations).
We also compare against the standard BERTScore and ROUGE metrics as well as a \qaeval{} variant which uses no answer verification by always marking the phrases as correct if the QA model predicts the question is answerable, denoted QA\-Eval-IsAns.

In general, all of the answer verification methods work comparably well, although BERTScore and LERC do statistically improve over the lexical methods in some settings, but not by large margins.
We believe the performance of \qaeval{}-Is\-Ans offers an explanation as follows.

Answer verification is not necessary if the QA model is perfect and the summaries are faithful (i.e., the QA prediction is always correct).
For \summeval{}, \citet{DeutschBeRo21} demonstrated that \qaeval{}'s QA performance was reasonable, and the summaries are very faithful with an average consistency score of 4.7 / 5 according to \citet{FKMSR21}.
Therefore, it may be difficult to demonstrate an improvement with any answer verification method even if it is high quality since the need for answer verification is low.
Indeed, we see \qaeval{}-Is\-Ans statistically ties the best methods.

On TAC'08, we expect it should be easier to show answer verification helps since \citet{DeutschBeRo21} showed the QA performance is poor, suggesting answer verification could help to suppress noisy predictions.
Indeed, we do see \qaeval{}-Is\-Ans is statistically out-performed by the verification methods.
We suspect the improvements are larger at the system-level than the summary-level because the system quality is estimated over a larger number of QA pairs than an individual summary's quality is.
A larger number of questions reduces any noise introduced by the verification methods, resulting in a more accurate estimate of summary quality and a better metric.

\begin{table}[t]
    \centering
    \small
    \begin{tabular}{lccccc}
        \toprule
        \multirow{2.4}{*}{\textbf{Metric}} & \multicolumn{2}{c}{\bf TAC'08} & & \multicolumn{2}{c}{\bf SummEval} \\
        \cmidrule{2-3} \cmidrule{5-6}
         & Sys & Sum & & Sys & Sum \\
         \midrule
BERTScore & \hphantom{$^\dagger$}.68$^\dagger$ & \hphantom{$^\dagger$}.40$^\dagger$ &  & \hphantom{$^\dagger$}.75$^\dagger$ & \hphantom{$^\dagger$}.27$^\dagger$ \\
ROUGE-1 & .60 & \hphantom{$^\dagger$}.39$^\dagger$ &  & .50 & .20 \\
ROUGE-2 & .67 & \hphantom{$^\dagger$}.39$^\dagger$ &  & .43 & .14 \\
\midrule
QAEval-IsAns & .63 & .37 &  & \hphantom{$^\dagger$}\underline{.70}$^\dagger$ & \hphantom{$^\dagger$}\bf .26$^\dagger$ \\
QAEval-EM & \hphantom{$^\dagger$}\bf .74$^\dagger$ & .29 &  & \hphantom{$^\dagger$}\underline{.77}$^\dagger$ & .19 \\
QAEval-F1 & .68 & .36 &  & \hphantom{$^\dagger$}\underline{.77}$^\dagger$ & .22 \\
QAEval-BERTScore & \hphantom{$^\dagger$}\underline{.68}$^\dagger$ & \hphantom{$^\dagger$}\underline{.38}$^\dagger$ &  & \hphantom{$^\dagger$}\underline{.77}$^\dagger$ & \hphantom{$^\dagger$}\bf .26$^\dagger$ \\
QAEval-LERC & \hphantom{$^\dagger$}\underline{.68}$^\dagger$ & \hphantom{$^\dagger$}\bf .39$^\dagger$ &  & \hphantom{$^\dagger$}\bf .80$^\dagger$ & \hphantom{$^\dagger$}\underline{.24}$^\dagger$ \\
        \bottomrule
    \end{tabular}
    \caption{System- and summary-level Kendall's $\tau$ (results with Pearson and Spearman are included in Appendix~\ref{sec:addtional_results}).
    Underlined QAEval values are statistically indistinguishable from the best QAEval scores (bottom) in bold.
    Values marked with $\dagger$ are statistically indistinguishable from the best metric overall (top and bottom).
    Statistical testing done using the single-tailed \textsc{Perm-Both} permutation test \citep{DeutschDrRo21} with $\alpha = 0.05$.
    }
    \label{tab:qaeval_correlations}
\end{table}
\begin{table}[t]
    \centering
    \small
    \begin{tabular}{lccc}
        \toprule
        \bf Metric & $r$ & $\rho$ & $\tau$ \\
        \midrule
ROUGE-1 & .13 & .13 & .11 \\
ROUGE-2 & .25 & .25 & .19 \\
BERTScore & .17 & .17 & .14 \\
FactCC & \hphantom{$^\dagger$}.34$^\dagger$ & \hphantom{$^\dagger$}.36$^\dagger$ & \hphantom{$^\dagger$}.29$^\dagger$ \\
FactCCX & .29 & .31 & .24 \\
\midrule
FEQA-EM & .17 & .14 & .11 \\
FEQA-F$_1$ & \bf .20 & \bf .16 & \bf .13 \\
FEQA-BERTScore & .15 & .12 & .10 \\
FEQA-LERC & .18 & \underline{.15} & \underline{.12} \\
        \bottomrule
    \end{tabular}
    \caption{The Pearson $r$, Spearman $\rho$, and Kendall $\tau$ correlations on the \summeval{} dataset.
    Values in bold are the best FEQA variants (bottom) with those underlined being statistically indistinguishable.
    $\dagger$ marks the best results across all metrics (top and bottom).}
    \label{tab:feqa_correlations}
\end{table}

\paragraph{FEQA}
We report the direct correlations between the human judgments and the FEQA variants, ROUGE, BERTScore, and FactCC \citep{KMXS20} in Table~\ref{tab:feqa_correlations}.
FactCC is a learned model to predict the factual consistency between two texts that was trained on synthetically generated data.

Among the FEQA variants, F$_1$ is the best or indistinguishable from LERC.
This result is expected given how similarly they perform at answer verification on this QA metric and dataset split.
This is again likely due to the fact that the summarization models copy heavily from the input documents, so the expected answers and QA model predictions are likely to be quite lexically similar.
Overall, the FEQA correlations are still lower than those by FactCC by a large margin.

It is also worth nothing that FEQA's correlations are lower than ROUGE-2's, a result which contradicts the findings of \citet{DurmusHeDi20}.
However, our experiments were conducted on a different dataset than theirs, and the two datasets' faithfulness scores were annotated in different ways.
Thus, we suspect the different conclusions are due different experimental setups;
the results cannot necessarily be fairly compared.

\section{Conclusion}
\label{sec:conclusion}
In this work, we benchmarked four different answer verification methods for QA-based summarization evaluation metrics.
Although we were able to identify that some methods perform better than others at verification, any such improvement does not necessarily translate a better overall metric quality.
We hypothesize that several factors, including the quality of the QA model and properties of the datasets, likely explain this result.
Even though token F$_1$ may be sufficient in some scenarios, we also recommend that practitioners also use LERC since it is likely to provide additional benefits, even if they are not easily measured.

\section*{Acknowledgments}
The authors would like to thank the anonymous reviews for their helpful suggestions, which we used to improve the final version of our work.

This research is supported by a Focused Award from Google and Contracts  FA8750-19-2-0201 and FA8750-19-2-1004 with the US Defense Advanced Research Projects Agency (DARPA).
Approved for Public Release, Distribution Unlimited.
The views expressed are those of the authors and do not reflect the official policy or position of the Department of Defense or the U.S. Government.

\bibliography{bibliography}
\bibliographystyle{acl_natbib}

\appendix

\section{Annotation Details}
\label{sec:annotation_details}

In total, 600 pairs of expected and predicted answers were annotated by one of the authors for whether or not they shared the same meaning.
The 600 pairs were sampled as follows:
QAEval was used to generate and predict questions on TAC'08 and Summ\-Eval and likewise for FEQA on Summ\-Eval.
Then, 200 questions were sampled uniformly at random from each metric and dataset combination.

The criteria for determining whether the two answers conveyed the same meaning was whether they could both be appropriately be used as synonyms given the input context and question.
In general, the annotation procedure was relatively straightforward with the majority of the answer pairs being clear synonyms of each other.
Example pairs are shown in Table~\ref{tab:examples}.
Some decisions did require world knowledge (e.g., ``Luis Enrique's side'' and ``Barcelona''), whereas others were clear synonyms (``EU'' and ``European Union'') or required resolving pronouns.
Decisions in cases which were not clear were based on the author's judgment of whether the two phrases seemed equally acceptable to use to answer the question, erring on the side of deciding the phrases are not semantically equivalent.
These cases were relatively uncommon.
\section{Additional Results}
\label{sec:addtional_results}
\begin{figure}
    \centering
    \includegraphics[width=0.95\columnwidth]{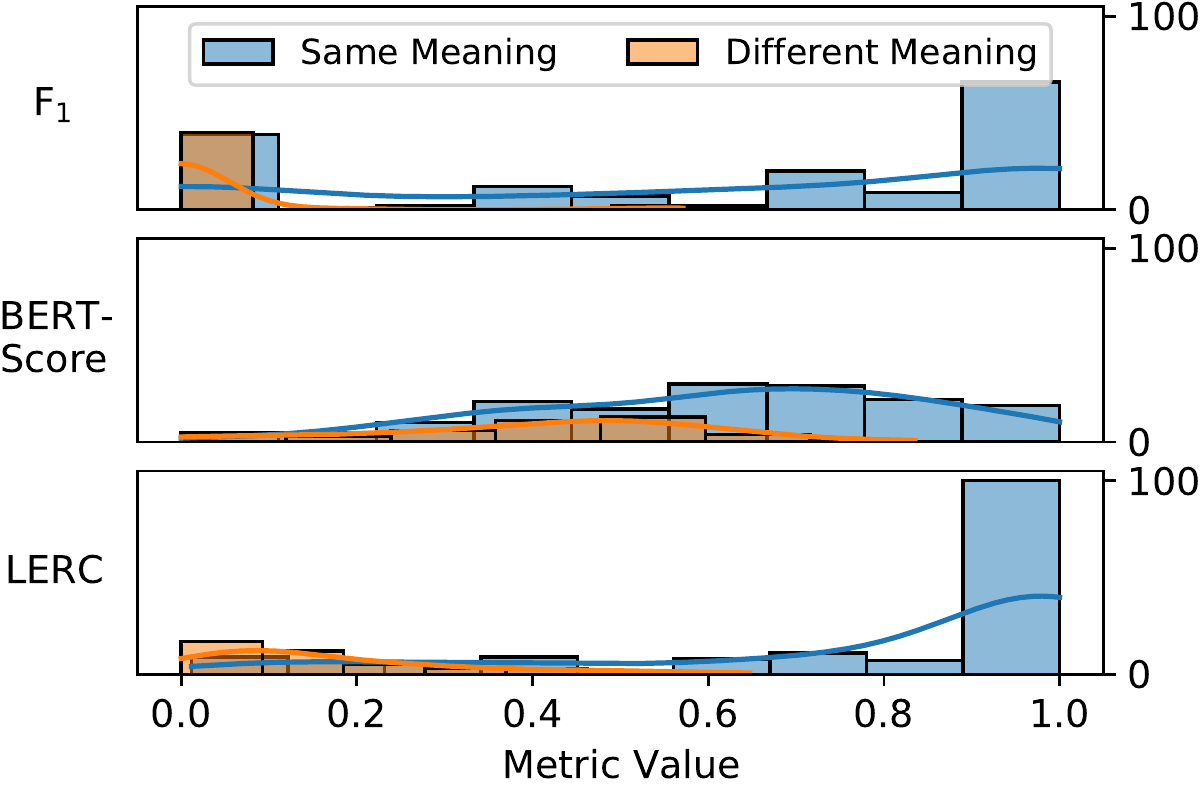}
    \caption{The distributions of score values for three metrics on the Summ\-Eval dataset for ground-truth answer and QA model prediction pairs from QAEval with the same (blue) and different (orange) meanings.}
    \label{fig:score_dist}
\end{figure}

Fig.~\ref{fig:score_dist} contains the distributions of score values for token F$_1$, BERTScore, and LERC on the \summeval{} dataset grouped by phrases that have and do no have the same meaning.
LERC most confidently separates the positive and negative examples.
F$_1$ performs similarly, except it fails in a large number of cases when the two phrases have no tokens in common.
BERTScore tends to mix the scores of the positive and negative classes, although they are separated on average.

\begin{table*}[t]
    \centering
    \begin{adjustbox}{width=\textwidth}
    \begin{tabular}{lccccccccccccccc}
    \toprule
    & \multicolumn{7}{c}{\bf TAC'08} & & \multicolumn{7}{c}{\bf SummEval}  \\
    \multirow{2}{*}{\textbf{Metric}} & \multicolumn{3}{c}{\textbf{System-Level}} & & \multicolumn{3}{c}{\textbf{Summary-Level}} & & \multicolumn{3}{c}{\textbf{System-Level}} & & \multicolumn{3}{c}{\textbf{Summary-Level}} \\
        \cmidrule{2-4} \cmidrule{6-8} \cmidrule{10-12} \cmidrule{14-16}
        & $r$ & $\rho$ & $\tau$ & & $r$ & $\rho$ & $\tau$ & & $r$ & $\rho$ & $\tau$ & & $r$ & $\rho$ & $\tau$ \\
        \midrule
BERTScore & .83 & \hphantom{$^\dagger$}.85$^\dagger$ & \hphantom{$^\dagger$}.68$^\dagger$ &  & \hphantom{$^\dagger$}.50$^\dagger$ & \hphantom{$^\dagger$}.50$^\dagger$ & \hphantom{$^\dagger$}.40$^\dagger$ &  & \hphantom{$^\dagger$}.84$^\dagger$ & \hphantom{$^\dagger$}.91$^\dagger$ & \hphantom{$^\dagger$}.75$^\dagger$ &  & \hphantom{$^\dagger$}.37$^\dagger$ & \hphantom{$^\dagger$}.35$^\dagger$ & \hphantom{$^\dagger$}.27$^\dagger$ \\
ROUGE-1 & .79 & .80 & .60 &  & \hphantom{$^\dagger$}.49$^\dagger$ & \hphantom{$^\dagger$}.48$^\dagger$ & \hphantom{$^\dagger$}.39$^\dagger$ &  & .61 & .62 & .50 &  & .28 & .26 & .20 \\
ROUGE-2 & .83 & \hphantom{$^\dagger$}.87$^\dagger$ & .67 &  & \hphantom{$^\dagger$}.48$^\dagger$ & \hphantom{$^\dagger$}.48$^\dagger$ & \hphantom{$^\dagger$}.39$^\dagger$ &  & .64 & .60 & .43 &  & .23 & .19 & .14 \\
ROUGE-L & .74 & .77 & .57 &  & .46 & .45 & .36 &  & .61 & .48 & .32 &  & .21 & .18 & .14 \\
ROUGE-SU4 & .80 & .83 & .63 &  & \hphantom{$^\dagger$}.49$^\dagger$ & \hphantom{$^\dagger$}.48$^\dagger$ & \hphantom{$^\dagger$}.39$^\dagger$ &  & .62 & .56 & .38 &  & .23 & .19 & .15 \\
QAEval-IsAns & .87 & .82 & .63 &  & \hphantom{$^\dagger$}.48$^\dagger$ & .47 & .37 &  & .76 & \hphantom{$^\dagger$}\underline{.86}$^\dagger$ & \hphantom{$^\dagger$}\underline{.70}$^\dagger$ &  & \hphantom{$^\dagger$}.33$^\dagger$ & \hphantom{$^\dagger$}\underline{.32}$^\dagger$ & \hphantom{$^\dagger$}\bf .26$^\dagger$ \\
QAEval-EM & \hphantom{$^\dagger$}\bf .92$^\dagger$ & \hphantom{$^\dagger$}\bf .89$^\dagger$ & \hphantom{$^\dagger$}\bf .74$^\dagger$ &  & .35 & .35 & .29 &  & \hphantom{$^\dagger$}\underline{.80}$^\dagger$ & \hphantom{$^\dagger$}\underline{.91}$^\dagger$ & \hphantom{$^\dagger$}\underline{.77}$^\dagger$ &  & .23 & .23 & .19 \\
QAEval-F1 & \hphantom{$^\dagger$}\underline{.90}$^\dagger$ & \hphantom{$^\dagger$}\underline{.86}$^\dagger$ & .68 &  & .46 & .45 & .36 &  & \hphantom{$^\dagger$}\underline{.82}$^\dagger$ & \hphantom{$^\dagger$}\underline{.91}$^\dagger$ & \hphantom{$^\dagger$}\underline{.77}$^\dagger$ &  & .30 & .29 & .22 \\
QAEval-BERTScore & \hphantom{$^\dagger$}\underline{.90}$^\dagger$ & \hphantom{$^\dagger$}\underline{.85}$^\dagger$ & \hphantom{$^\dagger$}\underline{.68}$^\dagger$ &  & \hphantom{$^\dagger$}\underline{.49}$^\dagger$ & \hphantom{$^\dagger$}\underline{.48}$^\dagger$ & \hphantom{$^\dagger$}\underline{.38}$^\dagger$ &  & \hphantom{$^\dagger$}\bf .84$^\dagger$ & \hphantom{$^\dagger$}\underline{.89}$^\dagger$ & \hphantom{$^\dagger$}\underline{.77}$^\dagger$ &  & \hphantom{$^\dagger$}\bf .36$^\dagger$ & \hphantom{$^\dagger$}\bf .34$^\dagger$ & \hphantom{$^\dagger$}\bf .26$^\dagger$ \\
QAEval-LERC & \hphantom{$^\dagger$}\underline{.89}$^\dagger$ & \hphantom{$^\dagger$}\underline{.85}$^\dagger$ & \hphantom{$^\dagger$}\underline{.68}$^\dagger$ &  & \hphantom{$^\dagger$}\bf .50$^\dagger$ & \hphantom{$^\dagger$}\bf .49$^\dagger$ & \hphantom{$^\dagger$}\bf .39$^\dagger$ &  & \hphantom{$^\dagger$}\underline{.81}$^\dagger$ & \hphantom{$^\dagger$}\bf .93$^\dagger$ & \hphantom{$^\dagger$}\bf .80$^\dagger$ &  & \hphantom{$^\dagger$}.33$^\dagger$ & \hphantom{$^\dagger$}\underline{.31}$^\dagger$ & \hphantom{$^\dagger$}\underline{.24}$^\dagger$ \\
    \bottomrule
    \end{tabular}
    \end{adjustbox}
    \caption{System- and summary-level correlations using Pearson's $r$, Spearman's $\rho$, and Kendall's $\tau$.}
    \label{tab:qaeval_all}
\end{table*}
\begin{table*}[t]
    \centering
    \begin{tabular}{llcc}
        \toprule
        \bf Answer & \bf Prediction & \bf \bf BERTScore & \bf LERC  \\
        \midrule
        EU & European Union & 0.73 & 0.84 \\ 
        a smaller leftist guerilla group & National Liberation Army & 0.48 & 0.10 \\
        six-time Olympic gold medalist & Usain Bolt & 0.34 & 0.35 \\
        Luis Enrique's side & Barcelona & 0.40 & 0.18 \\
        emergency responders & paramedics & 0.20 & 0.67 \\
        the child & toddler & 0.38 & 0.45 \\
        \bottomrule
    \end{tabular}
    \caption{
    Examples where BERTScore and LERC improve over F$_1$ (all examples have an F$_1$ score of 0).
    Successfully classing these phrases requires paraphrasing (e.g., ``the child'' and ``toddler'') and, in some cases, world knowledge (e.g., Usain Bolt had won six gold medals when the article was written).
    }
    \label{tab:examples}
\end{table*}

In Table~\ref{tab:qaeval_all}, we report the system- and summary-level correlations on TAC'08 and \summeval{} with Pearson's $r$ and Spearman's $\rho$ correlation coefficients in addition to the Kendall's $\tau$ which was presented in the main body of the paper.
The other coefficients lead to a similar conclusion to that which we made with Kendall's $\tau$:
All answer verification methods perform comparably well, and when BERTScore or LERC does improve over a lexical baseline, it is not by a large margin.
Further, using no verification method (\qaeval{}-Is\-Ans) largely performs equally well as \qaeval{} variants which do use a verification step on the \summeval{} dataset, but not on TAC'08.

\section{Example BERTScore/LERC Improvements}
\label{sec:examples}
Table~\ref{tab:examples} contains example expected answer and QA model prediction pairs for which BERTScore and LERC improve over exact match and token F$_1$.
We see that the improvements come from better identifying when the phrases are paraphrases of each other, which sometimes involves world knowledge.

\end{document}